\documentclass[runningheads]{llncs}
\usepackage[T1]{fontenc}
\usepackage{graphicx}
\usepackage{booktabs}
\usepackage{amsmath}
\usepackage{amssymb}
\usepackage{tabularx}
\usepackage{xspace}

\newcommand*{\lmss}{\fontfamily{lmss}\selectfont}

\newcommand{\PreStatic}{{\lmss Pre-Static$^\text{S}$}\xspace} 
\newcommand{\PreHipImg}{{\lmss Pre-Hip$^{\text{I}}$}\xspace} 
\newcommand{\PreChestImg}{{\lmss Pre-Chest$^{\text{I}}$}\xspace}
\newcommand{\PerVitals}{{\lmss Per-Vitals$^{\text{T}}$}\xspace} 
\newcommand{\Pre}{{\lmss Pre$^{\text{S+I}}$}\xspace} 
\newcommand{\Per}{{\lmss Per$^{\text{S+T}}$}\xspace} 
\newcommand{\All}{{\lmss All$^{\text{S+I+T}}$}\xspace}

\begin{document}
\title{Feature importance to explain multimodal prediction models. 
A clinical use case}

\titlerunning{Explainable Multimodal Mortality Prediction}

\author{
Jorn-Jan van de Beld\inst{1,2}\orcidID{0000-0001-6220-0508} \and
Shreyasi Pathak\inst{1}\orcidID{0000-0002-6984-8208} \and
Jeroen Geerdink\inst{2}\orcidID{0000-0001-6718-6653} \and
Johannes H. Hegeman\inst{1,2}\orcidID{0000-0003-2188-2738} \and
Christin Seifert\inst{3}\orcidID{0000-0002-6776-3868}
}

\authorrunning{J.J. van de Beld et al.}

\institute{
Faculty of EEMCS, University of Twente, 7500 AE Enschede, The Netherlands\\
\email{\{j.j.vandebeld,s.pathak\}@utwente.nl}\and
Hospital Group Twente (ZGT),  7609 PP Almelo, The Netherlands\\
\email{\{h.hegeman,j.geerdink\}@zgt.nl}\and
University of Marburg, 35037 Marburg, Germany\\
\email{christin.seifert@uni-marburg.de}
}
\maketitle  


\begin{abstract}
Surgery to treat elderly hip fracture patients may cause complications that can lead to early mortality.
An early warning system for complications could provoke clinicians to monitor high-risk patients more carefully and address potential complications early, or inform the patient. 
In this work, we develop a multimodal deep-learning model for post-operative mortality prediction using pre-operative and per-operative data from elderly hip fracture patients. Specifically, we include static patient data, hip and chest images before surgery in pre-operative data, vital signals, and medications administered during surgery in per-operative data. We extract features from image modalities using ResNet and from vital signals using LSTM.
Explainable model outcomes are essential for clinical applicability, therefore we compute Shapley values to explain the predictions of our multimodal black box model.
We find that i) Shapley values can be used to estimate the relative contribution of each modality both locally and globally, and ii) a modified version of the chain rule can be used to propagate Shapley values through a sequence of models supporting interpretable local explanations.
Our findings imply that a multimodal combination of black box models can be explained by propagating Shapley values through the model sequence.
\end{abstract}


\section{Introduction}
\label{sec:introduction}

The absolute number of hip fractures in the older Dutch population ($\geq$ 65 years) almost doubled between 1981 and 2008 from 7,614 to 16,049~\cite{phamPredictingHealthcareTrajectories2017}. In 1997, it was estimated that in 2050 4.5 million people worldwide will suffer from a hip fracture~\cite{gullbergWorldwideProjectionsHip1997}. Mortality is high during the early postoperative period with rates reported of up to 13.3\% in the first 30 days after surgery~\cite{huPreoperativePredictorsMortality2012}. 
An accurate risk score computed using data known before surgery \emph{(pre-operative data)} can lead to better information about the patient and be of help in the treatment selection~\cite{jonesRiskScoringSurgical1999}.
A risk score computed after surgery additionally using data from surgery \emph{(per-operative data)} could provide an early warning for complications, allowing for swift measures to mitigate the consequences.

In recent years, machine learning (ML) has proven to be promising for clinical practice to assist doctors in clinical decision-making and reduce workload~\cite{brigantiArtificialIntelligenceMedicine2020}. 
ML and deep neural network (DNN) models have been used to predict risks of certain complications after surgery~\cite{xueUseMachineLearning2021} and mortality using pre-operative and per-operative data~\cite{leeDevelopmentValidationDeep2018,fritzDeeplearningModelPredicting2019a}, reporting ROC-AUC scores up to 0.91.
Multimodal models use multiple data sources to predict outcomes, while unimodal models only consider a single data source. 
Clinicians use multiple data sources in their decision-making, therefore multimodal models should in theory be better at making these complex predictions~\cite{wang_big_2014}.

It is crucial that decisions made by an ML model can be understood by clinicians~\cite{tonekaboniWhatCliniciansWant2019}.
Some model types like decision trees are intrinsically explainable, however, more complex models trade performance at the cost of explainability~\cite{pawarExplainableAIHealthcare2020}. 
In this paper, we investigate to what extent explainability is hampered by using multimodal black box models that take multiple data sources (modalities) as input. 
We apply model agnostic explainability methods to understand the contribution of i)~each modality and ii)~individual features.

The contributions of our paper are as follows:
\begin{enumerate}
\item We present a neural model for predicting 30-day mortality of hip fracture patients combining three different data modalities (images, time-series, and structured data.

\item We apply XAI techniques, specifically Shapley values, to validate the mortality risk score given by our multimodal model.

\item We present the application of a novel method, specifically Shapley value propagation, to provide local explanations for a complex multimodal clinical prediction model.
\end{enumerate}


\section{Related Work}
\label{sec:relwork}

\begin{table*}[htp]
\label{tab:relwork_overview}

\caption{Overview of related work on short-term complication prediction. Abbreviations: convolutional neural network (CNN), auto-encoder (AE), random forest (RF), k-nearest neighbor (KNN), logistic regression (LR), support vector machine (SVM), gradient-boosted trees (GBT), decision tree (DT), naive bayes (NB), deep neural network (DNN), fully connected network (FC), long short-term memory (LSTM) network, XGBoost (XGB).}

\begin{tabularx}{\textwidth}{l X X X X}
\toprule
\textbf{Paper} & \textbf{Study\linebreak population} & \textbf{Prediction\linebreak target(s)} & \textbf{Data} & \textbf{ML model(s)}\textsuperscript{*} \\ \midrule
\cite{perngMortalityPredictionSeptic2019a} & Septic patients & Short-term mortality & Pre-operative static  & CNN, AE, RF, KNN, SVM \\
\cite{gowdConstructValidationMachine2019} & Total shoulder arthroplasty & Post-operative complications & Pre- and per-operative static  & LR, GBT, RF, KNN, DT, NB \\
\cite{schoenfeldAssessingUtilityClinical2016} & Spinal metastasis surgery & Short-term outcomes including mortality & Pre-operative static  & LR \\
\cite{leeDevelopmentValidationDeep2018} & Any surgery & Post-operative mortality & Pre- and per-operative static  & LR, DNN \\
\cite{fritzDeeplearningModelPredicting2019a} & Surgery with tracheal intubation & Post-operative short-term mortality & Pre- and per-operative and temporal & FC+LSTM+CNN \\
\cite{caoPredictiveValuesPreoperative2021} & Adult\textsuperscript{1} hip fracture patients & Post-operative short-term mortality & Pre-operative static  & LR, CNN \\
\cite{karresPredictingEarlyMortality2018} & Adult\textsuperscript{2} hip fracture patients & Post-operative short-term mortality & Pre-operative static  & LR \\
\cite{nijmeijerPredictionEarlyMortality2016} & Elderly\textsuperscript{3} hip fracture patients & Post-operative short-term mortality & Pre-operative static  & LR \\
\cite{yenidoganMultimodalMachineLearning2021} & Elderly\textsuperscript{3} hip fracture patients & Post-operative short-term mortality & Pre-operative static and images & LR, XGB, RF, SVM \\ \bottomrule
\end{tabularx}

\noindent{\footnotesize{\textsuperscript{1} Patients were at least 18 years old\\}}
\noindent{\footnotesize{\textsuperscript{2} Patients were at least 23 years old\\}}
\noindent{\footnotesize{\textsuperscript{3} Patients were at least 71 years old\\}}
\noindent{\footnotesize{\textsuperscript{*} Models separated with a comma were compared and models connected with a plus sign were combined}}

\end{table*}

In this section, we discuss the literature on three aspects; short-term complication prediction after surgery, multimodal prediction model, and explainability of ML models. 

\subsection{Short-term Complication Prediction}
ML models have been applied for short-term complication prediction after surgery, e.g. in septic~\cite{perngMortalityPredictionSeptic2019a} and hip fracture patients~\cite{caoPredictiveValuesPreoperative2021}. Table~\ref{tab:relwork_overview} summarises a sample of studies in literature featuring short-term complication prediction, showing pre-operative data as the most common input modality and logistic regression (LR) models as the most common ML model used in existing work.

Logistic regression models have been developed to predict early mortality ($<$30 days) after surgery in hip fracture patients using pre-operative static patient data, where reported AUC scores range from 0.76 to 0.82~\cite{nijmeijerPredictionEarlyMortality2016,caoPredictiveValuesPreoperative2021,karresPredictingEarlyMortality2018}.
In addition to pre-operative data, Yenidogan et al.~\cite{yenidoganMultimodalMachineLearning2021} also included pre-operative hip and chest images in their multimodal model. They extracted features from the images using convolutional neural networks (CNN) and trained a random forest (RF) to predict early mortality, where they reported an AUC of 0.79. However, the effect of per-operative data on risk score prediction after hip fracture surgery has not been investigated before.  

Cao et al. developed a model for the prediction of 30-day mortality of adult patients after hip fracture surgery using static data from 134,915 patients~\cite{caoPredictiveValuesPreoperative2021}. They compared the performance of a CNN with LR and reported an AUC of 0.76 for both models.
Yenidogan et al. addressed the 30-day mortality prediction after hip fracture surgery specifically for elderly patients~\cite{yenidoganMultimodalMachineLearning2021}. The authors exploited structured and image data available before surgery and showed significant improvement compared to their baseline the Almelo hip fracture score~(AHFS) developed by \cite{nijmeijerPredictionEarlyMortality2016}. 
They trained two CNNs to extract features from hip and chest X-ray images, which were fed to a RF classifier together with the structured data features. 
Their multimodal model (AUC=0.79) outperformed the AHFS baseline (AUC of 0.72).
Thus the authors concluded, that the additional information from multiple modalities is beneficial for model performance. In this paper, we add more modalities to further improve the prediction of postoperative mortality.

\subsection{Multimodal Prediction Models}
Clinical models that combine multiple modalities outperform models restricted to a single modality~\cite{demunterMortalityPredictionModels2017}. Multimodal models are commonly used for video classification tasks, where audio and image data are processed concurrently~\cite{nagraniAttentionBottlenecksMultimodal2021}.
The approaches differ in the way they share information between modalities within the neural network. Two common approaches are late and early fusion. Late fusion combines the predictions of the unimodal models with no further cross-modal information flow, while early fusion combines modalities very early in the pipeline~\cite{nagraniAttentionBottlenecksMultimodal2021}.

While early fusion allows for full information flow between modalities, it has a high computational cost, due to the high number of neuron connections. 
Late fusion has a low computational cost but does not enable the model to learn cross-modal interactions.

\subsection{Explainability}
Models are required to be explainable to gain the trust of clinicians, where knowing what features are most important to the model for its prediction is crucial~\cite{tonekaboniWhatCliniciansWant2019}.
Furthermore, clinicians need to be able to justify their decision-making towards patients and colleagues.
ML models can either be explained locally or globally~\cite{dasOpportunitiesChallengesExplainable2020}. Local explanations in a clinical setting focus on justifying the prediction for a single patient, while global explanations provide insight into general prediction tendencies for a larger population.

Multiple methods are available to compute the feature importance in deep learning models, most notably: LIME~\cite{ribeiroWhyShouldTrust2016}, deepLIFT~\cite{shrikumarLearningImportantFeatures2017}, layer-wise relevance propagation~\cite{bachPixelWiseExplanationsNonLinear2015} and Shapley values~\cite{lundbergUnifiedApproachInterpreting2017b}. In this paper, we use SHAP (SHapley Additive exPlanations) to estimate the importance of input features and modalities. 
Theoretically, all possible input feature subsets are considered during the computation of Shapley values, which makes them less sensitive to multicollinearity and thereby suitable for medical data.

Feature attribution methods, like SHAP, were not designed to explain complex multimodal clinical models~\cite{junaid_explainable_2023}. 
However, feature attributions can be propagated through a series models according to a modified version of the chain rule, which has been shown specifically for SHAP~\cite{chen_explaining_2022}. 
A multimodal model can be viewed as a sequence of unimodal models, which might make it suitable to be explained using the proposed chain rule.
In this study, we apply this method to provide local explanations for our multimodal clinical prediction model.


\section{Materials and Methods}
\label{sec:matmet}

In this section, we describe our dataset, our unimodal and multimodal models, our training procedure, and our evaluation.

\subsection{Dataset}
\label{subsec:dataset}
Our private data set contains anonymized data from 1669 patients who underwent hip fracture surgery at a public hospital between January 2013 and July 2021. 
Patients were at least 71 years old at the time of surgery. The data set contains 1189 (71.2\%) females and 480 males (28.8\%).
Our goal is to predict 30-day mortality after hip fracture surgery, which occurs at a rate of 7.8\% (131/1669) in our dataset.
We collected data from five modalities, which we divided into two groups - \textit{pre-operative} and \textit{per-operative} data.

\textit{Pre-operative data} encompasses information known before surgery, specifically: \textit{static patient data (Static)} modality, an \textit{axial hip X-ray image (HipImg)} modality and an \textit{anterior-posterior chest X-ray image (ChestImg)} modality. The static patient data has 76 features, which we further subdivided into seven categories: demographics, daily living conditions, nutrition, surgery information, lab results, medication, and comorbidities, for details see Appendix~\ref{apx:overview_features}.

\textit{Per-operative data} was collected during surgery containing \textit{vital signs (Vitals)} modality and \textit{medication data (Med)} modality.
The vital signs are heart rate, pulse, oxygen saturation, and blood pressure.
We split blood pressure into diastolic, systolic, and mean blood pressure resulting in a total of six time series.
The medication data includes 17 medication groups, which are commonly administered during hip fracture surgery, see Appendix~\ref{apx:overview_features}.
If a patient has missing medication data, then all values were set to zero, thus assuming the patient did not receive any medication during surgery at all.

\begin{figure*}[t]
\centering
\includegraphics[scale=0.4]{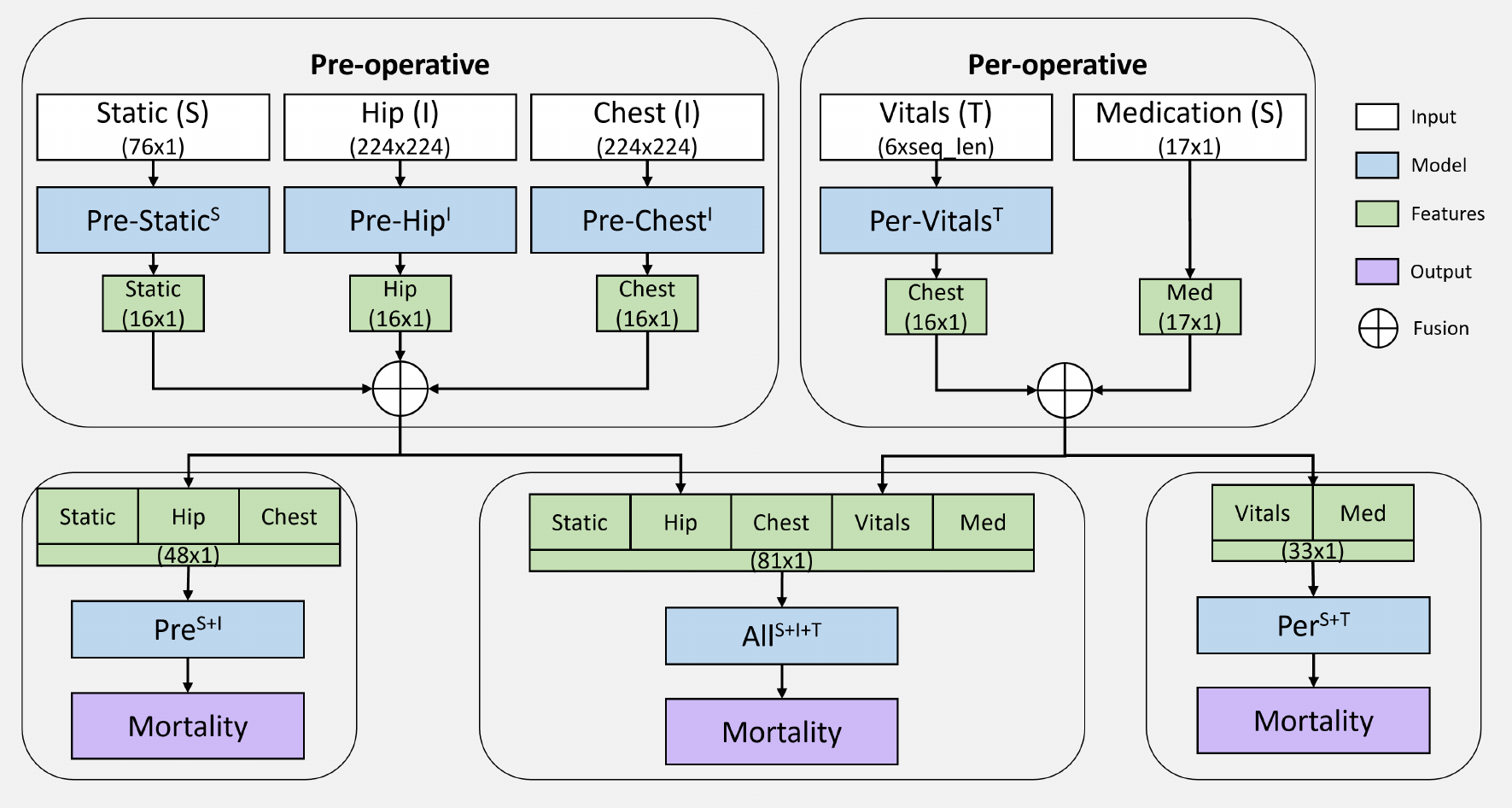}
\caption{Overview of how the unimodal models are fused to form the multimodal models. Dimensions at the input and feature extraction layers are shown in brackets, where the sequence length (\textit{seq\_len}) for the vitals varies between patients. No feature transformation was performed for the per-operative medication data. Data types: tabular/structured (S), image (I), time-series (T).}
\label{fig:multimodal_model_overview}
\end{figure*}

\subsubsection{Data preprocessing:} The pre-operative static patient data contain missing values, specifically 10 features had a missing percentage over 10\%. These included some the Charlson comorbidity index (66.6\%) and living situation (30\%). We imputed the data iteratively with a K-Nearest Neighbor Regression model\footnote{We used the scikit-learn implementation \url{https://scikit-learn.org/stable/modules/generated/sklearn.neighbors.KNeighborsRegressor.html}} ($k=10$).
We processed the vitals data, such that elements were spaced 15 seconds apart, where each element represents a single time step containing six vital signs. We filled up gaps of up to 5 minutes (20 elements) in each vital sign using linear interpolation. Furthermore, given the close similarity of heart rate and pulse, we interchangeably replaced missing values if only one value within the pair is missing.
So, if the heart rate was missing at a certain time step, then we took the pulse at that time step if available, and vice versa for missing pulse values.
We z-normalized the vital signs for each patient separately because this made fewer assumptions about the data population~\cite{karimInsightsLSTMFully2019}.
If after interpolation an element (time step) still contains any missing data, it is skipped during training and inference.

\subsection{Machine Learning Models}

We built the multimodal model (cf. Figure~\ref{fig:multimodal_model_overview}), by developing a model for each modality separately and fusing the representations of the single modalities. We trained the unimodal models to predict mortality. Then, we transferred the learned weights from unimodal models to the multimodal models to further train the latter. Table~\ref{tab:model_names} provides an overview of the models (abbreviation and description) that we consider in this paper. The remainder of this section details the unimodal models and the fusion approach to create the multimodal model.

\begin{table*}[htp]
\centering
\caption{Overview of models and their abbreviations. 
The per-operative vital signs have variable sequence length (\emph{len}). Data types: tabular/structured (S), image (I), time-series (T).}
\begin{tabularx}{\textwidth}{l X X l X X}
\toprule
\textbf{Model} & \textbf{Pre-\linebreak or\linebreak per-operative} & \textbf{Modality} & \textbf{Architecture} & \textbf{Input\linebreak size} & \textbf{Output\linebreak size} \\ \midrule
\PreStatic & pre & static patient data (S) & 1 FC & 76$\times$1 & 16$\times$1 \\
\PreHipImg & pre & hip (I) & ResNet50 & 224$\times$224 & 16$\times$1\\
\PreChestImg & pre & chest (I) & ResNet50 & 224$\times$224 & 16$\times$1 \\
\Pre & pre & static (S), hip (I), chest (I) & 2 FC & 48$\times$1 & 1 \\
\PerVitals & per & vital (T) & Bi-LSTM & 6$\times$\emph{len} & 16$\times$1 \\
\Per & per & vital (T), medication (S) & 2 FC & 33$\times$1 & 1\\
\All & both & static (S), hip (I), chest (I), vital (T), medication (S) & 2 FC & 81$\times$1 & 1 \\ \bottomrule
\end{tabularx}
\label{tab:model_names}
\end{table*}

\subsubsection{Pre-operative Models}
We developed three pre-operative unimodal models: the \PreStatic model for pre-operative static data and the \PreHipImg and \PreChestImg models for hip and chest images, respectively. 
For \PreStatic, the main task was dimensionality reduction, such that the pre-static feature dimension is reduced to the same number as the other modalities, before passing as input to the multimodal prediction model.
Therefore, we used a single fully connected hidden layer with 16 neurons with the leaky-ReLu activation function\footnote{
In the early stages of development, we observed that our models suffered from the ``dying ReLu" problem, therefore we chose to employ leaky-ReLu instead of the regular ReLu activation function for all our fully connected layers~\cite{agarapDeepLearningUsing2019}.}.   
For \PreHipImg and \PreChestImg, we used CNNs, which have emerged as the de-facto standard for medical image classification~\cite{caiReviewApplicationDeep2020}.
A wide range of CNN architectures are available, but given our small data set size, we chose ResNet50~\cite{heDeepResidualLearning2016}, a standard architecture with relatively few parameters for which a pre-trained model is available.
We added two fully connected layers with 256 and 16 neurons between ResNet50's global average pooling layer and the classification layer.
We used the same architecture for \PreHipImg and \PreChestImg.

\subsubsection{Per-operative Models}
The \PerVitals model takes 6 temporal features with variable sequence length as input, where the sequence lengths vary between 120 and 1337 elements, or 30 minutes to 5 hours, respectively.
We used bidirectional long short-term memory (Bi-LSTM) units~\cite{SchusterBidirectional1997} to extract information from the vital signs.
Our \PerVitals model contains a single Bi-LSTM layer with 2x128 units, followed by two hidden layers with 128 and 16 neurons. Timesteps still containing missing data after imputation were skipped during training and testing. Further, we encode the per-operative medication data with binary encoding.\footnote{In preliminary experiments we investigated ordinal and temporal encoding for the per-operative medication data but did not find a difference in performance.}
This data contained only 17 features, so contrary to the \PreStatic model we did not need to transform the input features to a lower-dimensional feature space.

\subsubsection{Multimodal Models}
We applied early fusion to generate three multimodal models: \Pre, \Per, and \All (see Table~\ref{tab:model_names}).
For each of these multimodal models, only the relevant modalities shown in Figure~\ref{fig:multimodal_model_overview} are included. We concatenated the pre-classification layer outputs from all unimodal models, where each modality supplies 16 features, except for the per-operative medication data, which yielded 17 features. 
The concatenation layer is followed by a fully connected layer with 64 neurons and a classification layer with a single neuron with sigmoid activation for mortality prediction.

\subsection{Training Procedure and Evaluation}
We randomly split our data into a training (50\%), validation (25\%), and test (25\%) set. We used repeated (N=5) cross-validation (k=5) to measure variability between training runs, furthermore, we used stratification to ensure a similar number of positive cases in each set. 
Models were optimized for maximum validation AUC, which was computed after each training epoch. 
We set the maximum number of epochs to 100 and used early stopping with a patience of 10. We halved the learning rate if there was no improvement for 5 epochs. We used the Adam optimizer and a batch size of 32. We tuned the learning rate of each model, by experimenting with a predefined range starting at $10^{-2}$ down to $10^{-5}$ with decrements of $10^{-1}$. The image models were trained with a relatively small learning rate of $10^{-5}$ because higher values led to low precision ($<$0.01).
The specific learning rates for each model can be found in Appendix \ref{apx:hyperparameters}.
To prevent overfitting, we added a dropout of 0.3 between fully connected layers and set the weight for L2 regularization to $10^{-3}$.

During training, our models were tasked with minimizing the weighted binary cross-entropy loss. 
The weight $w_{c_i}$ for class $c_i$ was computed according to Equation~\ref{eq:classweights} where $N_{total}$ is the total number of cases, $N_{c}=2$ is the number of classes and $N_{c_i}$ is the number of cases with class $c_i$~\cite{kingLogisticRegressionRare2001}.

\begin{equation}
    \label{eq:classweights}
    \frac{N_{total}}{N_c\cdot N_{c_i}}
\end{equation}

During the training of the image models, we augmented training images with random shift ($\pm$0.2), shear ($\pm$0.2), zoom ($\pm$0.2), and rotation ($\pm$20$^{\circ}$) to mimic a more diverse training set. We used bicubic interpolation to scale images to the 224x224 for ResNet50.

The multimodal models were initialized with the weights of the unimodal models, which we kept frozen while we trained the weights of the post-concatenation layers. 
Afterwards, we finetuned the model by unfreezing all layers, except for the layers in the image models. 
The learning rates were $5\cdot10^{-2}$ and $5\cdot10^{-3}$, for the first and second steps, respectively.

We computed recall, precision, F1-score, and AUC for the mortality prediction task. Each model architecture was trained 5 times with different randomly initialized weights, to measure variability between training runs. We report the mean and standard deviation of these runs.

The data are not publicly available due to privacy and ethical considerations, but all relevant code is available on GitHub\footnote{\url{https://github.com/jornjan/mmshap-XAI2024}}.

\subsection{Explanation}
We applied post-hoc model explanations to one version of the final model (\All) to better understand its predictions, thereby increasing clinical relevancy. To estimate the importance of features we used Shapley values\footnote{SHAP library \url{https://github.com/slundberg/shap}}, which resemble the contribution of each extracted feature towards a predicted outcome~\cite{lundbergUnifiedApproachInterpreting2017b}.
The contribution can be negative, indicating the value of the feature lowered the predicted value, or the contribution can be positive indicating the feature value increased the predicted value.

Given the multimodal nature of our model, we split the explanation into two steps.
In the first step, we compute the contribution of each modality towards the predicted outcome. In the second step, we compute the contribution of individual features from a given modality, specifically, we report the importance of individual pre-operative static features in the multimodal prediction, because we found that these contributed during our experiments.
In the first step, at the modality level, we provide local (single test case) and global (all test cases) explanations, while we limit ourselves to local explanations in the second step.

We define $H$ as the concatenation of the extracted hidden features from each modality ($H_{static}$, $H_{hip}$, $H_{chest}$ and $H_{vitals}$) and the per-operative medicine input values, $X_{med}$. 
We denote $f_{mm}$ as the function that maps the multimodal features in the concatenation layer to a 30-day mortality prediction ($\hat{y}$). The estimated Shapley values ($\hat{\phi}_{h_i}$) for each feature $h_i \in H$ have the property that they sum up to the predicted outcome~\cite{lundbergUnifiedApproachInterpreting2017b}.
$\hat{\phi}_0$ equals the average prediction based on the reference dataset, for which we took the training set.
\[
    H = H_{static} \oplus H_{hip} \oplus H_{chest} \oplus H_{vitals} \oplus X_{med}.
\]
\begin{equation}
    \hat{y} = f_{mm}(H) = \hat{\phi}_0 + \sum_{i=1}^{|H|} \hat{\phi}_{h_i}
\end{equation}

Global explanations follow from taking the average absolute contribution of each feature across all cases. We report global explanation for each modality, $m$ by calculating the absolute ($AC_{m}$) and relative ($RC_{m}$) contributions as follows:

\begin{align}
    AC_{m} &= \sum_{j=1}^{|H_{m}|} \hat{\phi}_{h_j} ,\mspace{20mu} h_j \in H_{m} \\
    RC_{m} &= \frac{AC_{m}}{\sum_{i=1}^{|H|} \hat{\phi}_{h_i}} ,\mspace{25mu} h_i \in H
\end{align}

We perform the second step, where we compute the contribution of individual static features for the multimodal prediction. 
Shapley values can be propagated through a sequence of models using a modified version of the chain rule~\cite{chen_explaining_2022}. Equation~\ref{eq:shapchain} shows how we computed the contribution of individual static features for a single case. We define $f_{static}$ as the function that extracts features from the static input features ($x_i$) and $AC_{static}$ as the absolute contribution of the extracted static features.

\begin{equation}
    \label{eq:shapchain}
    \hat{\phi}_{x_i} = \hat{\phi}(f_{static}, x_i) \mspace{5mu} (\hat{\phi}(f_{mm}, H_{static}) \mspace{5mu} \oslash \mspace{5mu} AC_{static}), \mspace{15mu} x_i \in X_{static}
\end{equation}


\section{Results}
\label{sec:results}

\subsection{Model Performance}

Table~\ref{tab:final_results} presents the average mortality prediction performance on the test set for all unimodal and multimodal models.
The \Pre, \PreStatic, and \All models all scored the highest average test set AUC of 0.78, which suggests adding modalities to the pre-operative static patient data does not improve performance.
Specifically, per-operative features seem to have low predictive power given the poor test set performances of the \PerVitals and \Per models.
All models tend to have a higher recall than precision, meaning the difficulty lies within correctly identifying non-risk patients. \emph{To summarize, pre-operative data has more predictive power than per-operative data. Specifically, the pre-operative static modality is the most important among all pre-operative modalities. This suggests that risk of mortality can be estimated pre-operatively by looking at static patient data and treatment options can be discussed prospectively.}

\begin{table}[ht]
\centering
\caption{Average performance of all models along with standard deviation over 25 runs. The best performance in each column is indicated in \textbf{bold}.}
\label{tab:final_results}
\begin{tabular}{@{}lllrrr@{}}
\toprule
\multicolumn{2}{c}{}                    & \multicolumn{1}{c}{AUC}  & \multicolumn{1}{c}{Recall} & \multicolumn{1}{c}{Precision} & \multicolumn{1}{c}{F1-score} \\ \midrule
\multicolumn{2}{l}{\textbf{Unimodal}}   &                          & \multicolumn{1}{l}{}       & \multicolumn{1}{l}{}          & \multicolumn{1}{l}{}         \\
      & \PreStatic       & $\textbf{0.78} \pm 0.04$ & $\textbf{0.82} \pm 0.09$   & $0.17 \pm 0.03$               & $0.28 \pm 0.04$              \\
      & \PreHipImg       & $0.50 \pm 0.07$          & $0.55 \pm 0.32$            & $0.09 \pm 0.04$               & $0.14 \pm 0.05$              \\
      & \PreChestImg     & $0.50 \pm 0.07$          & $0.55 \pm 0.29$            & $0.10 \pm 0.03$               & $0.15 \pm 0.03$              \\
      & \PerVitals       & $0.49 \pm 0.05$          & $0.49 \pm 0.30$            & $0.09 \pm 0.01$               & $0.15 \pm 0.02$              \\ \midrule

\multicolumn{2}{l}{\textbf{Multimodal}} &                          & \multicolumn{1}{l}{}       & \multicolumn{1}{l}{}          & \multicolumn{1}{l}{}         \\
      & \Pre             & $\textbf{0.78} \pm 0.04$ & $0.80 \pm 0.11$            & $\textbf{0.18} \pm 0.04$      & $\textbf{0.29} \pm 0.05$     \\
      & \Per             & $0.57 \pm 0.06$          & $0.62 \pm 0.23$            & $0.11 \pm 0.02$               & $0.17 \pm 0.02$              \\
      & \All             & $\textbf{0.78} \pm 0.04$ & $0.79 \pm 0.10$            & $0.17 \pm 0.02$               & $0.28 \pm 0.03$              \\ \bottomrule
\end{tabular}
\end{table}

\subsection{Explainability}

\textit{Global explanation:} Table~\ref{tab:modality_importance} shows the average relative contribution of each modality as well as the pre-operative and per-operative groups. 
The features extracted from the static patient data contribute the most with a relative contribution of 70.3\%, while other modalities contribute much less.
The second highest contributor is the per-operative vital signs with 10.4\% but with a high standard deviation of 10.5\%.
Also, the pre-operative image modalities have a standard deviation close to their mean value.
On average the per-operative medication data contributes the least towards the prediction.

\begin{figure}[p]\includegraphics[width=\textwidth]{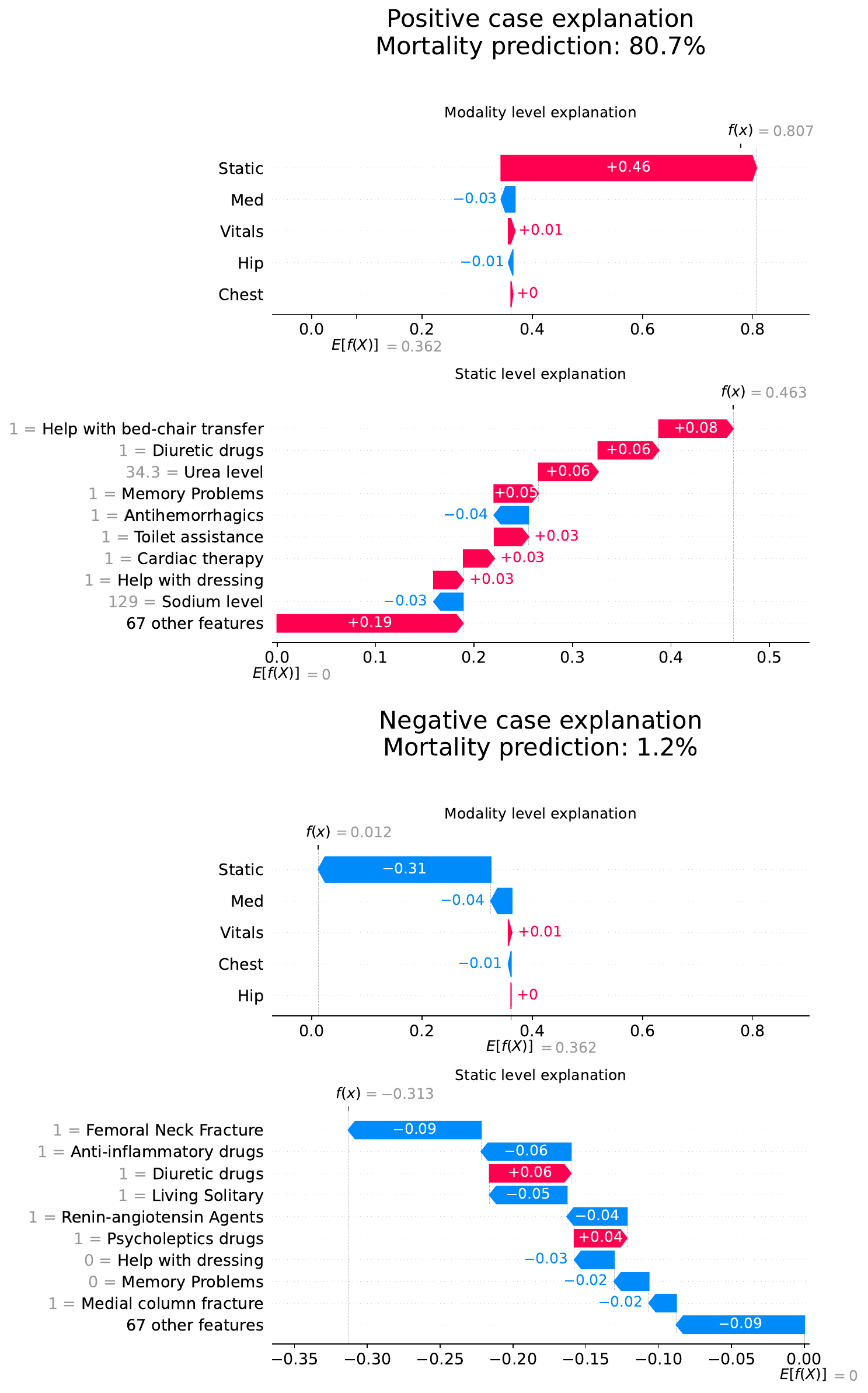}
    \caption{Shapley plots for single cases in the test set. First, the contribution of each modality is shown, followed by the contribution of individual static features.}
    \label{fig:local_shap}
\end{figure}

\textit{Local explanation:} Figure~\ref{fig:local_shap} shows a local explanation for two cases. 
Each case is explained with two plots, where the first shows how much each modality contributed towards the prediction, and the second shows feature-specific contribution for the \emph{static} modality.
Note that the contribution values of the \emph{Static} features sum up to the \emph{Static} contribution on the modality level.
The local multimodal explanations start at $\mathbb{E}[f(x)]=0.362$, which equals the average prediction for the training set. The plots show how each modality influenced the patient prediction compared to the training set average. 

The positive case prediction is strongly influenced by static features with a contribution of 46\% towards the prediction of 80.7\%, while the other modalities contribute 3\% or less.
The help with bed transfer is the most important static feature in this particular case with a contribution of 8\%. For the negative case prediction (plots at the bottom), the static features are most important in reducing the mortality prediction by 31\%. 
The fact that this case concerned a femoral neck fracture reduced the prediction by 9\%.

\begin{table}
\centering
\small
\caption{Mean relative contribution for mortality prediction of input modalities taken over 25 runs and in brackets the standard deviation. RC: relative contribution.}
\begin{tabular}{@{}lllr@{}}
\toprule
\multicolumn{2}{l}{\textbf{Modality}} & \multicolumn{2}{l}{\textbf{RC \% (st. dev.)}} \\ \midrule
\multicolumn{2}{l}{\textbf{Pre-operative}} & \multicolumn{2}{c}{\textbf{84.6\% (11.1)}} \\
 & Static patient data &  & 70.3\% (10.6)\\
 & Hip image &  & 7.0\% (6.0) \\
 & Chest image &  & 7.3\% (8.3) \\ \midrule
\multicolumn{2}{l}{\textbf{Per-operative}} & \multicolumn{2}{c}{\textbf{15.4\% (11.1)}} \\
 & Vitals &  & 10.4\% (10.5) \\
 & Medication &  & 5.0\% (2.0) \\ \bottomrule
\end{tabular}
\label{tab:modality_importance}
\end{table}

\section{Discussion}
\label{sec:discussion}

\emph{Comparison to state-of-the-art:} The addition of per-operative data did not yield a significant performance improvement compared to our pre-operative multimodal model.
Our unimodal image models (\PreHipImg and \PreChestImg) score worse compared to the image models developed by Yenidogan et al.~\cite{yenidoganMultimodalMachineLearning2021}, who reported an AUC of 0.67 and 0.70 on the hip and chest images, respectively. 
The lower performance could be due to our smaller data set and smaller CNN architecture.
On the other hand, our \PreStatic model performs on par with state-of-the-art models with a smaller dataset~\cite{yenidoganMultimodalMachineLearning2021,nijmeijerPredictionEarlyMortality2016,caoPredictiveValuesPreoperative2021}.

\emph{Usage of explanations:} Using Shapley values, we explained the prediction of the \All model at the modality level.
Globally the pre-operative static features are most important with a relative importance of 70.3\%, still the relative importance of per-operative features in our \All model was 15.4\%. Further, we provide an example of a local explanation in Figure~\ref{fig:local_shap}.
We explained our multimodal model in two steps. First, the contribution of the extracted features from each modality was computed with which we calculated the contribution of individual pre-operative static features in the second step. 
We only show the explanation for the pre-operative static data as we found this to be the most important modality. Explanations for other unimodal models can be added as well, e.g. using Grad-CAM~\cite{selvaraju2017grad} for images. 
Static features describing kidney function (Urea level, diuretic drug, and sodium level) are important for both the single-patient local explanations for predicting 30-day mortality.
Furthermore, the level of self-reliance is important given the importance of: help with bed-chair transfer, help with dressing, and living situation.

\emph{Limitations and future work:} Our results show that predicting mortality for hip fracture patients is difficult  
and starting with \emph{predicting} if any \emph{complication} occurs within 30 days could improve the assessment of the data set predictive capabilities. 
Coincidentally, this resolves the class imbalance issue, because in our data set 49\% of the patients experience at least one complication within 30 days after surgery. 
Furthermore, as an intermediate step complications could be grouped by severity or cases could be scored on a scale from no complication to mortality. It would be interesting to look into this in future work.
Clinically, this could help determine, whether a patient is at risk after surgery and requires more attention.

Our \emph{fusion method} could be described as mid fusion, because we used the pre-classification layer of each unimodal model, however we did do dimension reduction before concatenation. This method ensured each modality contributed the same number of features to the classification layer, however, this might not be optimal for post-operative complication prediction. 
Future work could include a deeper investigation of fusion methods, like late fusion and bottleneck fusion.
Bottleneck fusion restricts cross-modal information flow by using a very limited amount of neurons for information to pass through. The idea is that the model is forced to condense the most important cross-modal features leading to better performance at negligible computational cost~\cite{nagraniAttentionBottlenecksMultimodal2021}.

During our experiments, we found that the models that include static patient data or vital signs are overfitting the data suggesting that the model might be memorizing the training data, with poor \emph{generalization capability} to the test data. This might indicate that: i) there is some variation within the instances and does not contain many common patterns, ii) there is not enough data for generalization. 
The addition of dropout layers and L2-regularization did not solve this problem, therefore a different approach is required.
It has been shown that reducing the number of pre-operative static features based on their importance can prevent overfitting~\cite{caoPredictiveValuesPreoperative2021}.
We could use the Shapley values to iteratively select the most important feature, up until we reach a certain subset size.
Additionally, if we prevent strong overfitting on the pre-operative static data, this could incentivize the multimodal models to focus more on the other modalities for information.

Our multimodal model is not robust against \emph{missing data}, only the per-operative medication data and part of the pre-operative static data are allowed to be missing. In clinical practice, this would mean patients are excluded if pre-operative images or per-operative vitals are missing. We impute the pre-operative static patient data, so having some missing values there does not lead to exclusion, however, the imputation quality is dependent on the number of non-missing features. Therefore, for clinical applicability future models should be robust against \emph{missing modalities}, to avoid patient exclusion.

\section{Conclusion}
\label{sec:conclusion}


We conclude that the addition of per-operative data to pre-operative data does not significantly improve 30-day mortality prediction.
Further, investigation confirmed that pre-operative features are most important for mortality prediction, while per-operative features contribute little except for a few per-operative medications.
We show that multimodal black box models can be made explainable by applying the chain rule to the feature attributions of each model in the sequence.
Future work, could restrict to only using pre-operative data and further explain model predictions by providing interpretable explanations for the contribution of the image modalities.

\subsubsection{Disclosure of Interests.}
Christin Seifert is a member of the XAI 2024 World Conference committee. All other authors have no competing interests to declare that are relevant to the content of this article.
%


\bibliographystyle{splncs04.bst}
\bibliography{references.bib}


\appendix
\section{Hyperparameters}
\label{apx:hyperparameters}
\begin{table}[ht]
\caption{Hyperparameters for each model: Learning Rate (LR)}
\centering
\label{tab:hyperparameters}
\begin{tabular}{@{}lr@{}}
\toprule
\textbf{Model}  & \textbf{LR}  \\ \midrule
\PreStatic   & $10^{-3}$ \\
\PreHipImg    & $10^{-5}$ \\
\PreChestImg  & $10^{-5}$ \\
\PerVitals & $5\cdot 10^{-4}$ \\
\Pre    & $5\cdot 10^{-2}$ \\
\Per    & $5\cdot 10^{-2}$ \\
\All    & $5\cdot 10^{-2}$  \\ \bottomrule
\end{tabular}
\end{table}

\section{Detailed feature overview}
\label{apx:overview_features}

\begin{table*}[ht]
\centering
\caption{Available pre-operative features grouped by category}
\label{tab:preop_features}
\resizebox{\textwidth}{!}{%
\begin{tabular}{@{}llll@{}}
\toprule
\textbf{Demographics} & \textbf{\begin{tabular}[c]{@{}l@{}}Daily living\\ activities\end{tabular}} & \textbf{Nutrition} & \textbf{\begin{tabular}[c]{@{}l@{}}Surgery\\ information\end{tabular}} \\ \midrule
Age & Help with transfer from bed to chair & Malnutrition risk & Fracture type \\
Surgery start/end & Help with showering & Unintended weight loss & Surgery type \\
Falling risk & Help with dressing & Drink or tube feeding & Fracture laterality \\
Fall last year & Help with going to toilet & Decreased appetite & ASA score \\
Pre-fracture mobility & Help with eating & SNAQ score &  \\
Living situation & Help with self-care &  &  \\
Prone to delirium & Katz ADL score &  &  \\
Memory problems & Incontinence material used &  &  \\
Delirium in the past &  &  &  \\
CCI score &  &  &  \\ \bottomrule
\end{tabular}%
}
\end{table*}

\begin{table*}[ht]
\centering
\begin{tabular}{@{}lll@{}}
\toprule
\textbf{Lab results} & \textbf{Medication (reason/effect)} & \textbf{Comorbidities} \\ \midrule
HB & Blood thinners & Chronic pulmonary disease \\
HT & Vitamin D & Congestive heart failure \\
CRP & Polypharmacy & Peripheral vascular disease \\
LEUC & A02 (acid related disorders) & Cerebrovascular disease \\
THR & A10 (diabetes) & Dementia \\
BLGR & B01 (antithrombotic) & Renal disease \\
IRAI & B02 (antihemmorrhagics) & Rheumatological disease \\
ALKF & B03 (antianemic) & Cancer \\
GGT & C01 (cardiac therapy) & Cerebrovascular event \\
ASAT & C03 (diuretics) & Liver disease \\
ALAT & C07 (beta blockers) & Lymphoma \\
LDH1 & C08 (calcium channel blockers) & Leukemia \\
UREU & C09 (renin-angiotensin system) & Peptic ulcer disease \\
KREA & C10 (lipid modification) & Diabetes \\
GFRM & L04 (immunosuppressants) & Prior myocardial infarction \\
NA & M01 (anti-inflammatory) &  \\
XKA & N05 (psycholeptics) &  \\
GLUCGLUC & R03 (airway obstruction) &  \\ \bottomrule
\end{tabular}

\end{table*}

\begin{table*}[ht]
\centering
\caption{List of medications that were at least administered in 100 unique cases, also includes the general reason for usage. Percentages are with respect to a total of 1616 cases for which medication data is available.}
\begin{tabular}{@{}lrr@{}}
\toprule
\textbf{Medication} & \multicolumn{1}{l}{\textbf{\# unique cases}} & \textbf{Effect} \\ \midrule
Bupivacaine & 649 (40.2\%)& Anesthetic \\
Cefazolin & 692 (42.8\%) & Antibiotic \\
Dexemethasone & 198 (12.3\%) & Anti-inflammatory \\
Efedrine & 445 (27.5\%) & Increase blood pressure \\
Elektrolytes & 468 (29.0\%) & Minerals \\
Esketamine & 561 (34.7\%) & Anesthetic \\
Lidocaine & 405 (25.1\%) & Anesthetic \\
Metamizole & 187 (11.6\%) & Painkiller \\
Midazolam & 475 (29.4\%) & Anesthetic \\
Noradrenaline & 887 (54.9\%) & Increase blood pressure \\
Ondansetron & 282 (17.5\%) & Counter post-operative nausea \\
Piritramide & 446 (27.6\%) & Painkiller \\
Propofol & 648 (40.1\%) & Anesthetic \\
Rocuronium & 260 (16.1\%) & Muscle relaxant \\
Sufentanil & 746 (46.2\%) & Painkiller \\
Sugammadex & 112 (6.9\%) & Reverse muscle relaxant \\
Tranexamic acid & 465 (28.8\%) & Prevent blood loss \\ \bottomrule
\end{tabular}%
\label{tab:medication_list}
\end{table*}

\end{document}